\documentclass[conference,letterpaper]{IEEEtran}
\IEEEoverridecommandlockouts
\IEEEsettextwidth{0.76in}{0.76in}
\IEEEsetsidemargin{c}{0pt}
\IEEEsettextheight{1.03in}{0.76in}
\IEEEsettopmargin{t}{1.03in}
\usepackage{cite}
\usepackage{amsmath,amssymb,amsfonts}
\usepackage{algorithm}
\usepackage{algorithmic}
\usepackage{graphicx}
\usepackage{textcomp}
\usepackage{xcolor}
\usepackage{url}
\def\BibTeX{{\rm B\kern-.05em{\sc i\kern-.025em b}\kern-.08em
    T\kern-.1667em\lower.7ex\hbox{E}\kern-.125emX}}

\begin{document}

\title{Scalable Bayesian Optimization for Efficient Exploration of Moir\'e Patterns
\thanks{This work was supported by the European Union's Horizon Europe research and innovation programme under grant agreement No. 101135927 (NOUS).}
}

\author{\IEEEauthorblockN{1\textsuperscript{st} Panagiotis Krokidas}
\IEEEauthorblockA{\textit{Institute of Informatics and}\\
\textit{Telecommunications}\\
\textit{NCSR ``Demokritos''}\\
Agia Paraskevi, Greece \\
p.krokidas@iit.demokritos.gr}
\and
\IEEEauthorblockN{2\textsuperscript{nd} Christos Lazaridis}
\IEEEauthorblockA{\textit{Department of Informatics} \\
\textit{University of Piraeus}\\
Piraeus, Greece \\
p22083@unipi.gr}
\and
\IEEEauthorblockN{3\textsuperscript{rd} Vassilis Sioros}
\IEEEauthorblockA{\textit{Department of Informatics and Telecommunications} \\
\textit{National and Kapodistrian University of Athens}\\
Athens, Greece; vsioros@di.uoa.gr\\[1pt]
\textit{Institute of Informatics and Telecommunications} \\
\textit{NCSR ``Demokritos''}\\
Agia Paraskevi, Greece; v.sioros@iit.demokritos.gr}
\and
\IEEEauthorblockN{4\textsuperscript{th} Efi-Maria Papia}
\IEEEauthorblockA{\textit{Institute of Nanoscience and} \\ \textit{Nanotechnology} \\
\textit{NCSR ``Demokritos''}\\
Agia Paraskevi, Greece \\
e.papia@inn.demokritos.gr}
\and
\IEEEauthorblockN{5\textsuperscript{th} Vassilios Constantoudis}
\IEEEauthorblockA{\textit{Institute of Nanoscience and} \\ \textit{Nanotechnology} \\
\textit{NCSR ``Demokritos''}\\
Agia Paraskevi, Greece \\
v.constantoudis@inn.demokritos.gr}
\and
\IEEEauthorblockN{6\textsuperscript{th} George Giannakopoulos}
\IEEEauthorblockA{\textit{Institute of Informatics and}\\ \textit{Telecommunications} \\
\textit{NCSR ``Demokritos''}\\
Agia Paraskevi, Greece \\
ggianna@iit.demokritos.gr}
}

\maketitle

\begin{abstract}
Large materials design spaces often contain rare high-performing configurations that are difficult to identify through trial-and-error exploration or purely representative machine-learning (ML) sampling. In this work, we present FruBO, a flexible Bayesian Optimization framework for accelerating the discovery of optimal configurations in computationally demanding materials design problems. As a case study, we investigate multilayer Moir\'e patterns generated by the superposition of six rotated hexagonal lattices, where the objective is to identify the rotational conformation that maximizes pore-size uniformity, quantified using the Jain fairness index. The resulting benchmark contains 600,000 unique configurations and exhibits a highly skewed objective distribution, making the global optimum difficult to locate by random sampling. We compare random exploration, Gaussian Process-based Bayesian Optimization, and Bayesian Optimization using a lightweight CatBoost surrogate with committee-based uncertainty estimation. Random sampling identifies the global optimum only after 533,012 evaluations, requiring approximately 31 hours, while Gaussian Process-based BO becomes impractical due to GPU memory exhaustion after 290 acquired samples. In contrast, FruBO with CatBoost identifies the global optimum after 1,470 acquired samples and approximately 1.15 hours. These results demonstrate that scalable surrogate models can substantially improve the feasibility and efficiency of BO in large nanomaterials design spaces.
\end{abstract}


\section{Introduction}
\label{sec:intro}
Materials science, and particularly the development of advanced materials, inherently involves navigating vast design spaces in search of rare optimal candidates that often lie far from the average behavior of most sampled configurations. As a result, identifying high-performing materials rapidly exceeds the capabilities of traditional trial-and-error approaches, even when assisted by domain expertise and intuition refined through continuous experimentation. Consequently, Machine Learning (ML) has attracted significant interest across nanomaterials research, where predictive models are trained on representative subsets of sampled configurations to estimate material properties and performance. However, such representative sampling primarily captures average trends, while accurately identifying rare, exceptional, or highly optimal cases remains unlikely~\cite{Krokidas2025}. Therefore, conventional ML approaches alone are often insufficient for efficiently discovering the rare high-performing solutions of greatest scientific and technological interest. To address this limitation, inverse design and AI-driven optimization methods have emerged as powerful approaches for navigating complex design spaces toward optimal solutions. These include evolutionary strategies, particle swarm optimization~\cite{Tani2021,Tani2024,Santoni2024}, and Bayesian Optimization (BO)~\cite{shahriari2016bo}. Among them, BO is widely regarded as one of the most effective frameworks due to its ability to guide sampling through a surrogate model coupled with an acquisition function balancing exploration and exploitation. In materials science and chemistry, BO has successfully accelerated searches toward high-performing regions in applications including material and drug discovery~\cite{Chitturi2024,Wu2024,Krokidas2025}. Nevertheless, standard BO pipelines commonly rely on Gaussian Processes (GPs) as surrogate models, whose computational and memory requirements scale poorly with the number of acquired samples $N$, typically approaching $\mathcal{O}(N^3)$ complexity and $\mathcal{O}(N^2)$ memory usage~\cite{Rasmussen2006,Siemenn2023_}. Consequently, although BO is often more sample-efficient than evolutionary or swarm-based approaches~\cite{Tani2021,Tani2024,Santoni2024}, GP-related overhead can eventually dominate the optimization workflow, particularly for large-scale problems and modest computational infrastructures.

In this work, we present FruBO, a flexible and extensible Bayesian Optimization library designed to address the computational limitations of conventional BO pipelines while improving optimization performance. FruBO incorporates a broad range of functionalities and optimization strategies; in the present study, we focus on enabling the use of alternative surrogate models that substantially reduce computational overhead compared to traditional GP-based approaches. Importantly, these lightweight surrogates are not only computationally more efficient, but can also achieve superior optimization performance across challenging optimization tasks. For this study, we use Moir\'e patterns as a testbed for evaluating our approach, enabling the construction of a controlled benchmark representative of complex nanomaterials design spaces. Our results show that properly selected surrogate models can overcome GP limitations while achieving high computational efficiency and strong optimization performance.

\section{Moir\'e Optimization Problem}

\subsection{Moir\'e Patterns and Pore Uniformity}

Moir\'e patterns arise when periodic meshes or lattices are superimposed with slight differences in orientation or spacing, producing larger-scale interference structures composed of regions of local alignment and misalignment \cite{andrei2021marvels,chen2020moire,shi2021exotic}. In materials science, these structures have attracted considerable interest because they provide a controllable mechanism for tuning material properties \cite{musaddaq2020,PAPIA2026114270}. In layered van der Waals materials such as graphene and transition-metal dichalcogenides, small rotational offsets generate Moir\'e superlattices capable of significantly modifying the electronic band structure \cite{xu2021creation}. Such effects have driven the emergence of twistronics, where rotational alignment is exploited to engineer electronic, optical, and magnetic properties for applications in nanoelectronics, photonics, and quantum technologies \cite{jadaun2023review}. Beyond electronic behavior, Moir\'e structures are increasingly important for porous and functional nanomaterials, since the relative orientation between layers directly affects the resulting pore-size distribution. These effects are relevant in applications such as filtration membranes, photonic devices, and engineered magnetic materials \cite{pryds2024twisted,hu2021tailoring,song2021direct}. For example, multilayer porous graphene studies have demonstrated that rotational alignment can substantially influence water permeability and transport properties \cite{park2022effect}. In this work, we focus on the optimization of pore-size uniformity in multilayer Moir\'e structures. Uniform pore distributions are desirable because they promote homogeneous transport behavior. To quantify pore uniformity, we employ the Jain fairness index, a metric commonly used to evaluate the uniformity of a distribution. For a set of pore areas $\{x_i\}_{i=1}^{n}$, the Jain fairness index is defined as:

\begin{equation}
J(\mathbf{x}) =
\frac{\left(\sum_{i=1}^{n} x_i\right)^2}
{n \sum_{i=1}^{n} x_i^2}
\end{equation}

where values approaching 1 correspond to highly uniform pore distributions, while lower values indicate increasingly heterogeneous pore sizes. The optimization objective of this study is therefore to identify rotational configurations that maximize the Jain fairness index.

\begin{figure}[htbp]
\centering
\includegraphics[width=0.8\linewidth]{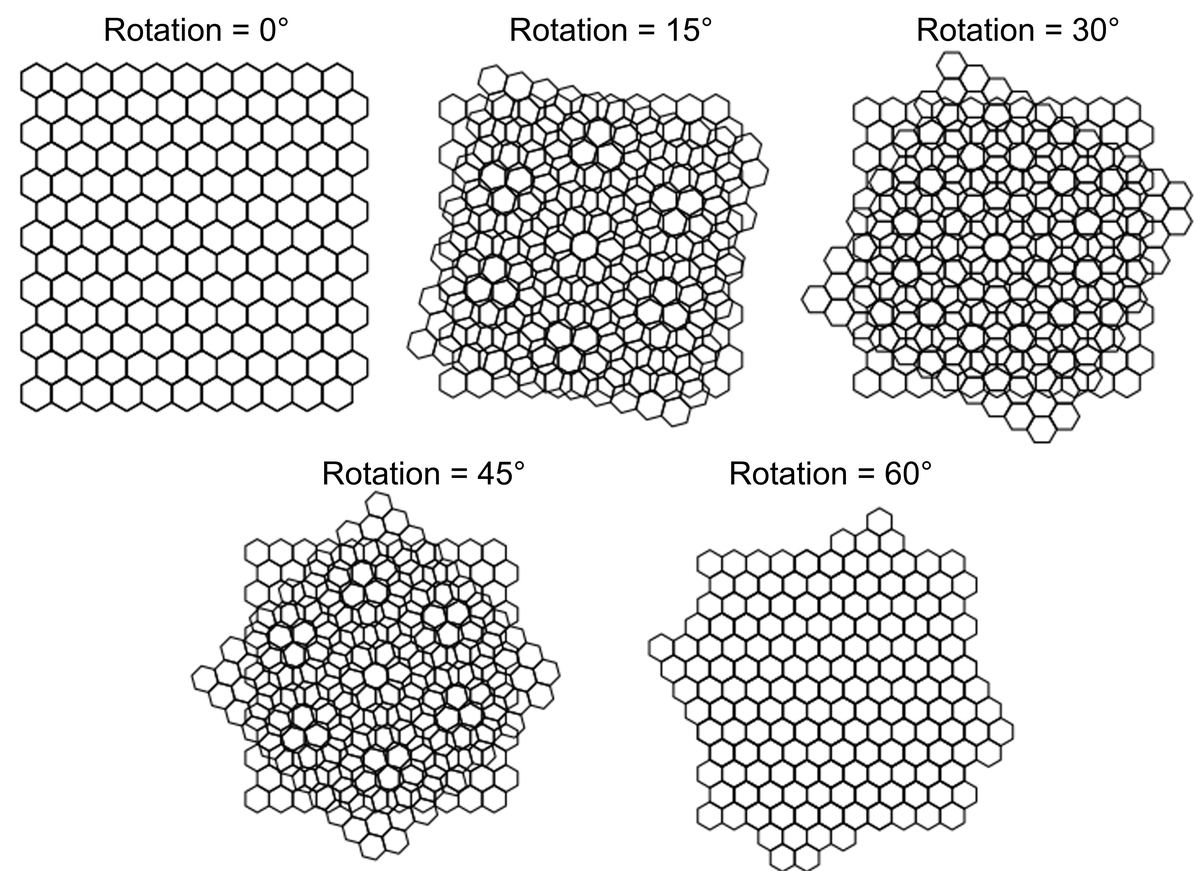}
\caption{Illustration of Moir\'e pattern formation generated from the superposition of two hexagonal lattices under different relative rotational offsets. Even small angular variations produce substantially different interference geometries and pore distributions, demonstrating the strong dependence of the resulting structure on layer orientation.}
\label{moire}
\end{figure}

Figure~\ref{moire} illustrates the effect of rotational offsets on overlapping hexagonal lattices. Even relatively small angular changes produce substantially different interference patterns and pore geometries.

\subsection{Creating the Moir\'e Benchmark Dataset}
\label{sec:dataset}

To construct the optimization problem, we generate synthetic multilayer Moir\'e configurations by stacking six hexagonal mesh layers with independently controlled rotational angles. Each configuration corresponds to a unique combination of rotational parameters defining the relative orientation of the layers. The geometry generation process is based on periodic hexagonal ring structures arranged on a two-dimensional lattice, where a specific parameter controls the number of hexagonal lattice elements generated for each layer and therefore determines the spatial extent and geometric resolution of the resulting Moir\'e structure. For every sampled rotational configuration, the overlapping layers produce a distinct pore arrangement. The resulting pore regions are identified computationally, and their geometric characteristics are extracted in order to calculate the corresponding Jain fairness index. This establishes a direct mapping between the rotational configuration parameters and a scalar objective value describing pore-size uniformity. The final dataset consists of 600,000 unique six-layer rotational configurations, forming a large discrete design space with high combinatorial complexity. Evaluating the entire design space becomes computationally demanding due to both the geometric generation process and the pore-distribution analysis required for each sample. Consequently, the problem provides a realistic and challenging benchmark for evaluating scalable and resource-efficient Bayesian Optimization approaches for advanced materials design.

\begin{figure}[htbp]
\centering
\includegraphics[width=0.8\linewidth, trim=0 10 0 5,
  clip]{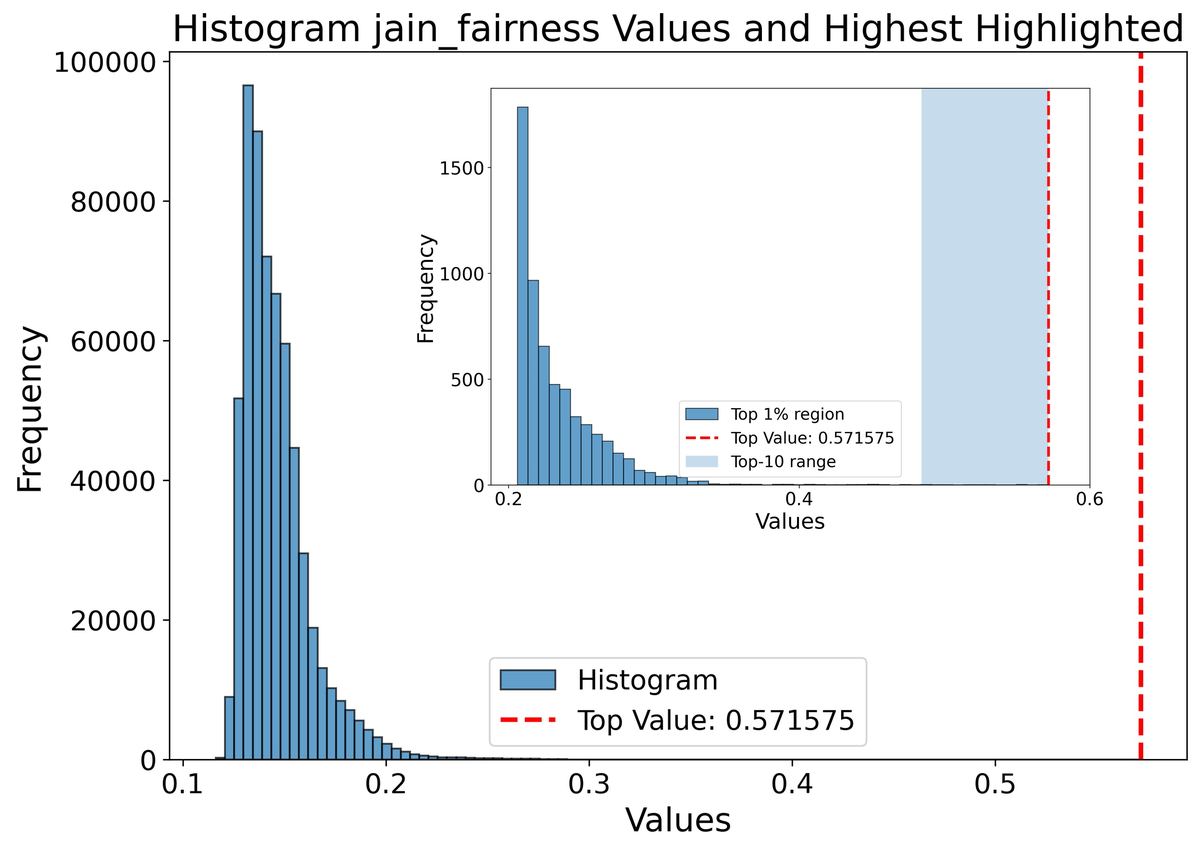}
\caption{Distribution of the Jain fairness index across the 600,000-sample Moir\'e dataset. The inset shows a zoomed view of the high-value tail, highlighting the extreme rarity and isolation of the globally optimal configurations, which makes their identification particularly challenging for random or exhaustive sampling approaches.}
\label{distribution}
\end{figure}

\section{The Scientific Premise}
Scientific discovery in large design spaces often requires identifying an optimal configuration among a vast number of candidates, where each configuration corresponds to a different performance outcome. In materials science, such optimization tasks commonly arise in both \textit{in vivo} and \textit{in silico} experimentation, where evaluating each candidate may involve expensive laboratory procedures, computational simulations, or geometric analyses. Traditionally, exploration follows a trial-and-error methodology based on intuition, prior experience, or random sampling. However, as the dimensionality and combinatorial complexity of the design space grow, exhaustive evaluation rapidly becomes computationally prohibitive, while unguided exploration may require a very large number of evaluations before approaching an optimal solution. In this work, we investigate this challenge in the context of multilayer Moir\'e structures, where the objective is to identify rotational configurations that maximize pore-size uniformity, quantified through the Jain fairness index. To address this problem, we compare two exploration strategies: random sampling, representing the traditional trial-and-error paradigm, and Bayesian Optimization (BO), where surrogate models and acquisition functions guide sampling toward promising regions of the search space while minimizing the number of expensive evaluations. The following paragraphs describe the two exploration strategies evaluated in this work.

\textbf{Trial-and-error approach:}
The first approach represents the traditional trial-and-error paradigm through random sampling of the design space. Configurations are selected uniformly from a search space of $M$ possible Moir\'e conformations. Each sampled configuration requires generation of the corresponding geometry, computation of the pore-size distribution, and evaluation of the Jain fairness index. Sampling proceeds iteratively in batches of $N$ evaluations, while the best objective value found so far is monitored until improvements reach a plateau.

\textbf{Bayesian Optimization for guided sampling:}
The second approach replaces unguided exploration with Bayesian Optimization (BO), where sampling is guided by an iteratively updated surrogate model. An initial set of $N$ randomly selected configurations is evaluated and used to train a model that approximates the relationship between the rotational angles and the Jain fairness index. The surrogate then predicts the performance of unevaluated configurations and provides the predictive mean ($\mu$) and uncertainty ($\sigma$), which are used by an acquisition function to rank candidate samples. The highest-ranked batch of configurations is evaluated, added to the training set, and the surrogate is retrained. Repeating this closed-loop process progressively improves the model's knowledge of the design space and guides sampling toward promising regions while balancing exploration and exploitation. Common acquisition functions include Expected Improvement (EI), Upper Confidence Bound (UCB), and Probability of Improvement (PI)~\cite{Wang2023}. As discussed in Section~\ref{sec:intro}, conventional BO typically relies on Gaussian Processes (GPs), whose computational and memory requirements become prohibitive as the number of acquired samples increases. In this work, these limitations motivate the investigation of alternative surrogate models that reduce computational overhead while maintaining or improving optimization performance.

\section{BO specifics -- acquisition function and surrogate models}
In this work, BO is driven by the \textit{q}Log Expected Improvement (qLogEI) acquisition function, a batch variant of the Expected Improvement (EI) criterion, with 10 points sampled per iteration. For every unevaluated configuration, the surrogate model provides two quantities: the predicted objective value (mean, $\mu$) and the associated predictive uncertainty (standard deviation, $\sigma$). qLogEI combines these quantities to estimate the expected improvement over the best solution identified so far, prioritizing configurations that are predicted to perform well, exhibit high uncertainty, or both. Unlike the sequential EI formulation, qLogEI enables the simultaneous selection of multiple candidate configurations. In our experiments, a batch size of ten configurations is acquired at each optimization iteration. As discussed in Section~\ref{sec:intro}, conventional BO pipelines predominantly rely on Gaussian Processes (GPs) to provide the predictive mean and uncertainty required by the acquisition function. However, the computational and memory overhead of GP models can become prohibitive as the number of acquired samples increases. In the present work, these limitations become evident in the Moir\'e optimization problem, motivating the investigation of alternative surrogate models that not only alleviate the computational overhead of GP-based BO, but also achieve superior optimization performance. Among the models already implemented in FruBO, we evaluated CatBoost~\cite{Prokhorenkova2018}. Since CatBoost does not inherently estimate predictive uncertainty, uncertainty was approximated using a query-by-committee strategy comprising five independently trained CatBoost models initialized with different random seeds. The CatBoost hyperparameters used in the surrogate model were \texttt{iterations}=300, \texttt{depth}=6, \texttt{learning\_rate}=0.03, and \texttt{loss\_function}=\texttt{RMSE}.

\section{The Experiments}
As described in Section~\ref{sec:dataset}, we consider 600,000 configurations in six dimensions, corresponding to six layers. For both random sampling and BO, the results are obtained from 10 independent runs with different seeds.


\begin{figure}[!htbp]
\centering
\includegraphics[height=0.42\textheight,keepaspectratio, trim=0 14 0 5,
  clip]{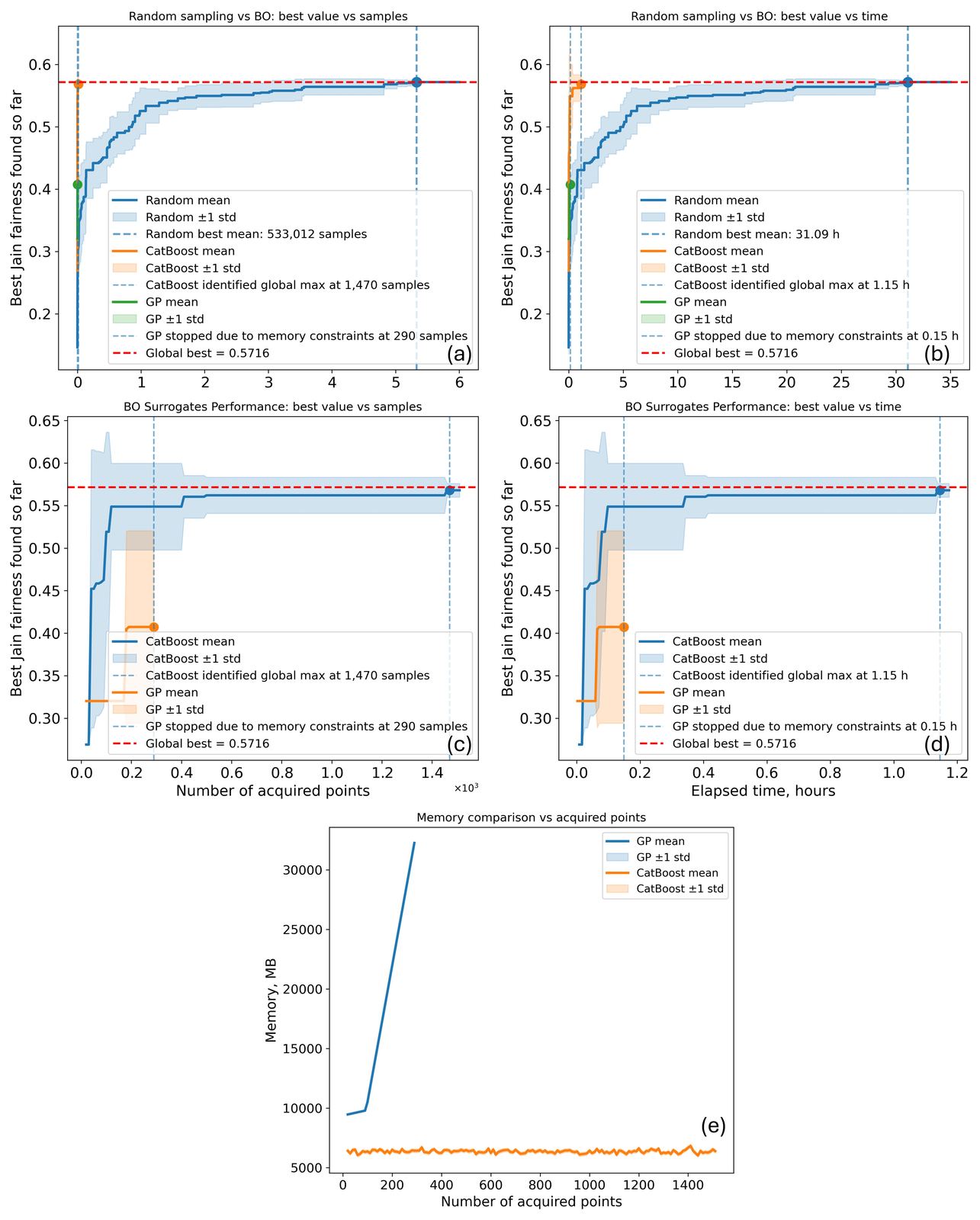}
\caption{Optimization performance on the Moir\'e dataset. (a)--(b) Best Jain fairness index found versus acquired samples and elapsed time for random sampling and Bayesian Optimization (BO) using GP and CatBoost surrogate models. (c)--(d) Zoomed comparison of the BO surrogates. (e) Memory consumption of the surrogate models during optimization.}
\label{results}
\end{figure}

The results of the trial-and-error approach shown in Fig.~\ref{results}(a)--(b) demonstrate that, on average, the optimal solution (corresponding to the six-layer conformation with the highest Jain fairness index, which is 0.5716) is identified only after 533,012 sampled configurations, requiring approximately 31 hours of computation. In contrast, applying BO with the traditional Gaussian Process (GP) (specifically the baseline GP from GPyTorch \cite{botorch}) surrogate proved impractical, as the steadily increasing memory consumption exhausted the available GPU resources after only 290 acquired samples (Fig.~\ref{results}(e)). This behavior confirms the computational limitations of GP-based BO discussed in Section~\ref{sec:intro}.

To overcome this limitation, we replaced GP with CatBoost as a surrogate model. The results (Fig.~\ref{results}(e)) show that CatBoost is free from the memory bottleneck affecting GPs, providing a lightweight and computationally efficient surrogate for large-scale BO. Using CatBoost, BO successfully identified the global optimum (Jain fairness index = 0.5716) after acquiring only 1,470 samples, corresponding to approximately 1.15 hours of computation on our hardware. Compared to random sampling, this represents a reduction of more than two orders of magnitude in the number of required evaluations and approximately a 27-fold reduction in wall-clock time. Table~\ref{tab:optimization_summary} summarizes the main performance differences among the evaluated strategies.

\begin{table}[htbp]
\caption{Summary of optimization performance on the Moir\'e dataset}
\begin{center}
\begin{tabular}{|l|c|c|c|}
\hline
\textbf{Method} & \textbf{Samples} & \textbf{Time (h)} & \textbf{Outcome} \\
\hline
Random sampling & 533,012 & 31.00 & Optimum found \\
\hline
BO--GP & 290 & -- & GPU memory exhausted \\
\hline
BO--CatBoost & 1,470 & 1.15 & Optimum found \\
\hline
\end{tabular}
\label{tab:optimization_summary}
\end{center}
\end{table}

\section{Conclusions}
We presented FruBO, a flexible Bayesian Optimization framework for exploring large materials design spaces. The 600,000-configuration Moir\'e benchmark highlighted the difficulty of finding rare optimal configurations through unguided sampling. This benchmark provides a demanding test of both sample efficiency and computational feasibility under limited resources. GP-based BO exhausted the available GPU memory early in the optimization process. In contrast, the CatBoost surrogate implemented through FruBO avoided this memory bottleneck while preserving the predictive mean and uncertainty estimates required by the acquisition function. Using this lightweight surrogate, BO identified the global optimum with 1,470 acquired samples and approximately 1.15 hours of computation, compared with 533,012 samples and approximately 31 hours for random sampling. These findings indicate that replacing standard GP surrogates with appropriate ML models can substantially improve the scalability of BO for complex nanomaterials optimization problems.

\section{Compute Resources}
BO experiments were executed with the in-house-developed software package FruBO (\url{https://github.com/insane-group/FruBO/}). All calculations, including the creation of the Moir\'e dataset, calculation of the Jain fairness index for each conformation, and the BO runs, were carried out on a Windows desktop equipped with an Intel Core i9 processor, 64~GB of RAM, and an NVIDIA RTX 3070 Ti GPU.


\begin{thebibliography}{00}
\bibitem{Krokidas2025} P. Krokidas, V. Gkatsis, J. Theocharis, and G. Giannakopoulos, ``Navigating materials design spaces with efficient Bayesian optimization: a case study in functionalized nanoporous materials,'' \textit{Digital Discovery}, vol. 4, pp. 3753--3763, 2025, doi: 10.1039/d5dd00237k.
\bibitem{Tani2021} L. Tani, D. Rand, C. Veelken, and M. Kadastik, ``Evolutionary algorithms for hyperparameter optimization in machine learning for application in high energy physics,'' \textit{European Physical Journal C}, vol. 81, no. 2, pp. 1--9, 2021, doi: 10.1140/epjc/s10052-021-08950-y.
\bibitem{Tani2024} L. Tani and C. Veelken, ``Comparison of Bayesian and particle swarm algorithms for hyperparameter optimisation in machine learning applications in high energy physics,'' \textit{Computer Physics Communications}, vol. 294, p. 108955, 2024, doi: 10.1016/j.cpc.2023.108955.
\bibitem{Santoni2024} M. L. Santoni, E. Raponi, R. De Leone, and C. Doerr, ``Comparison of high-dimensional Bayesian optimization algorithms on BBOB,'' \textit{ACM Transactions on Evolutionary Learning and Optimization}, vol. 4, no. 3, 2024, doi: 10.1145/3670683.
\bibitem{shahriari2016bo} B. Shahriari, K. Swersky, Z. Wang, R. P. Adams, and N. De Freitas, ``Taking the human out of the loop: A review of Bayesian optimization,'' \textit{Proceedings of the IEEE}, vol. 104, no. 1, pp. 148--175, 2016, doi: 10.1109/JPROC.2015.2494218.
\bibitem{Wang2023} X. Wang, Y. Jin, S. Schmitt, and M. Olhofer, ``Recent advances in Bayesian optimization,'' \textit{ACM Computing Surveys}, vol. 55, no. 13s, 2023, doi: 10.1145/3582078.
\bibitem{Chitturi2024} S. R. Chitturi \textit{et al.}, ``Targeted materials discovery using Bayesian algorithm execution,'' \textit{npj Computational Materials}, vol. 10, no. 1, pp. 1--12, 2024, doi: 10.1038/s41524-024-01326-2.
\bibitem{Wu2024} Y. Wu, A. Walsh, and A. M. Ganose, ``Race to the bottom: Bayesian optimisation for chemical problems,'' \textit{Digital Discovery}, vol. 3, no. 6, pp. 1086--1100, 2024, doi: 10.1039/d3dd00234a.
\bibitem{Rasmussen2006} C. E. Rasmussen and C. K. I. Williams, \textit{Gaussian Processes for Machine Learning}. Cambridge, MA, USA: MIT Press, 2006.
\bibitem{Siemenn2023_} A. E. Siemenn and T. Buonassisi, ``Decreasing the computing time of Bayesian optimization using generalizable memory pruning,'' in \textit{Proc. 2023 IEEE High Performance Extreme Computing Conf. (HPEC)}, 2023, doi: 10.1109/HPEC58863.2023.10363486.
\bibitem{andrei2021marvels} E. Y. Andrei \textit{et al.}, ``The marvels of moir\'e materials,'' \textit{Nature Reviews Materials}, vol. 6, pp. 201--206, 2021, doi: 10.1038/s41578-021-00284-1.
\bibitem{chen2020moire} X. Chen \textit{et al.}, ``Moir\'e engineering of electronic phenomena in correlated oxides,'' \textit{Nature Physics}, vol. 16, no. 6, pp. 631--635, 2020.
\bibitem{song2021direct} T. Song \textit{et al.}, ``Direct visualization of magnetic domains and moir\'e magnetism in twisted 2D magnets,'' \textit{Science}, vol. 374, no. 6571, pp. 1140--1144, 2021.
\bibitem{du2023moire} L. Du, M. R. Molas, Z. Huang, G. Zhang, F. Wang, and Z. Sun, ``Moir\'e photonics and optoelectronics,'' \textit{Science}, vol. 379, no. 6639, p. eadg0014, 2023.
\bibitem{pryds2024twisted} N. Pryds, D. S. Park, T. S. Jespersen, and S. Yun, ``Twisted oxide membranes: A perspective,'' \textit{APL Materials}, vol. 12, no. 1, 2024.
\bibitem{hu2021tailoring} G. Hu, M. Wang, Y. Mazor, C.-W. Qiu, and A. Al\`u, ``Tailoring light with layered and moir\'e metasurfaces,'' \textit{Trends in Chemistry}, vol. 3, no. 5, pp. 342--358, 2021.
\bibitem{shi2021exotic} B. Shi, P. Qi, M. Jiang, Y. Dai, F. Lin, H. Zhang, and Z. Fang, ``Exotic physical properties of 2D materials modulated by moir\'e superlattices,'' \textit{Materials Advances}, vol. 2, no. 17, pp. 5542--5559, 2021.
\bibitem{xu2021creation} Y. Xu \textit{et al.}, ``Creation of moir\'e bands in a monolayer semiconductor by spatially periodic dielectric screening,'' \textit{Nature Materials}, vol. 20, no. 5, pp. 645--649, 2021.
\bibitem{jadaun2023review} P. Jadaun and B. Sore\'e, ``Review of orbital magnetism in graphene-based moir\'e materials,'' \textit{Magnetism}, vol. 3, no. 3, pp. 245--258, 2023.
\bibitem{park2022effect} C. Park, F. Robinson, and D. Kim, ``Effect of layer orientation and pore morphology on water transport in multilayered porous graphene,'' \textit{Micromachines}, vol. 13, no. 10, p. 1786, 2022, doi: 10.3390/mi13101786.
\bibitem{PAPIA2026114270} E.-M. Papia, A. Kondi, D. Nioras, F. Sofos, and V. Constantoudis, ``Data-driven investigation of analytical relationships in moir\'e structured materials,'' \textit{Computational Materials Science}, vol. 261, p. 114270, 2026, doi: 10.1016/j.commatsci.2025.114270.
\bibitem{musaddaq2020} M. Azeem \textit{et al.}, ``Optimal design of multilayer fog collectors,'' \textit{ACS Applied Materials \& Interfaces}, vol. 12, no. 6, pp. 7736--7743, 2020.
\bibitem{Lakshminarayanan2017} B. Lakshminarayanan, A. Pritzel, and C. Blundell, ``Simple and scalable predictive uncertainty estimation using deep ensembles,'' in \textit{Proc. 31st Int. Conf. Neural Information Processing Systems (NIPS)}, 2017, pp. 6405--6416.
\bibitem{Prokhorenkova2018} L. Prokhorenkova, G. Gusev, A. Vorobev, A. V. Dorogush, and A. Gulin, ``CatBoost: Unbiased boosting with categorical features,'' \textit{Advances in Neural Information Processing Systems}, vol. 2018-Decem, pp. 6638--6648, 2018.
\bibitem{botorch} M. Balandat \textit{et al.}, ``BoTorch: A framework for efficient Monte-Carlo Bayesian optimization,'' in \textit{Advances in Neural Information Processing Systems}, 2020, arXiv:1910.06403.
\end{thebibliography}
\end{document}